\def\year{2022}\relax
\documentclass[letterpaper]{article} % DO NOT CHANGE THIS
\usepackage{aaai22}  % DO NOT CHANGE THIS
\usepackage{times}  % DO NOT CHANGE THIS
\usepackage{helvet}  % DO NOT CHANGE THIS
\usepackage{courier}  % DO NOT CHANGE THIS
\usepackage[hyphens]{url}  % DO NOT CHANGE THIS
\usepackage{graphicx} % DO NOT CHANGE THIS
\usepackage{natbib}  % DO NOT CHANGE THIS AND DO NOT ADD ANY OPTIONS TO IT
\usepackage{caption} % DO NOT CHANGE THIS AND DO NOT ADD ANY OPTIONS TO IT
\DeclareCaptionStyle{ruled}{labelfont=normalfont,labelsep=colon,strut=off} % DO NOT CHANGE THIS
\usepackage{epsfig}
\usepackage{amsmath}
\usepackage{amssymb}
\usepackage{dsfont}
\usepackage{subfigure}
\usepackage[ruled,vlined]{algorithm2e}

\usepackage{multirow}
\usepackage{booktabs}

\SetKwInput{KwInput}{Input}
\SetKwInput{KwOutput}{Output}

\DeclareMathOperator*{\argmin}{arg\,min}

\usepackage[pagebackref=true,breaklinks=true,letterpaper=true,colorlinks,bookmarks=false]{hyperref}

\newcommand{\ie}{\textit{i}.\textit{e}.}
\newcommand{\eg}{\textit{e}.\textit{g}.}

\usepackage{newfloat}
\usepackage{listings}

\title{Proxyless Neural Architecture Adaptation for Supervised Learning and Self-Supervised Learning}
\author{
    Do-Guk Kim\equalcontrib\thanks{Corresponding authors}\textsuperscript{\rm 1}, Heung-Chang Lee\equalcontrib$^\dagger$\textsuperscript{\rm 2}
}
\affiliations{
    \textsuperscript{\rm 1}Department of Artificial Intelligence, Inha University\\
    dgkim@inha.ac.kr\\
    \textsuperscript{\rm 2}AI Lab, Kakao Enterprise\\ andrew.com@kakaoenterprise.com
}

\usepackage{bibentry}
\begin{document}

\maketitle

\begin{abstract}
Recently, Neural Architecture Search (NAS) methods are introduced and show impressive performance on many benchmarks.
Among those NAS studies, Neural Architecture Transformer (NAT) aims to adapt the given neural architecture to improve performance while maintaining computational costs.
However, NAT lacks reproducibility and it requires an additional architecture adaptation process before network weight training.
In this paper, we propose proxyless neural architecture adaptation that is reproducible and efficient.
Our method can be applied to both supervised learning and self-supervised learning.
The proposed method shows stable performance on various architectures.
Extensive reproducibility experiments on two datasets, \ie, CIFAR-10 and Tiny Imagenet, present that the proposed method definitely outperforms NAT and be applicable to other models and datasets.
\end{abstract}

%%%%%%%%% BODY TEXT
\section{Introduction}
\label{Intro}
Traditionally, neural architectures are manually designed by human experts: \eg VGG, ResNet, and DenseNet.
Choosing proper neural architecture for a given dataset and task is important for getting high performance.
Recently, architectures selected by Neural Architecture Search (NAS) algorithms achieved state-of-the-art performances on various benchmark datasets.
Even though NAS methods show great performance on various tasks and datasets, their prohibitively high computational cost makes NAS methods hard to be applied practically.

To overcome this limitation, many recent works focused on reducing the computational costs of NAS while maintaining the advantage of NAS approaches.
Among them, several studies attempted to adapt given architectures for the given dataset and task rather than finding architectures within the vast search space.
Neural Architecture Transformer (NAT)~\cite{guo2019nat} is one of the representative works on network architecture adaptation.
The authors attempted to improve the given architecture so that the performance of the network is improved while maintaining or reducing the computational costs.
However, NAT still requires an architecture adaptation stage in addition to the conventional network weight training stage.
To make architecture adaptation effective and practical, we need to improve architecture during the network training stage.

Also, NAT has several other limitations.
First, the reproducibility of the algorithm is not verified since the authors reported only one result for each model.
Second, it can only transform the neural networks with identical cell architectures.
Recent NAS works focus on searching macroblock-based architectures that have various cell architectures.
However, those architectures cannot be transformed by NAT.

In this paper, we propose an architecture adaptation technique that doesn't require an additional architecture adaptation stage.
Since the proposed method improves architectures during the conventional train stage, it is \textit{proxyless} and can be easily combined with the existing deep neural network training process.
Especially, we examined the effectiveness of the proposed method not only for supervised learning but also self-supervised representation learning.

Our contributions are summarized as:
\begin{itemize}
    \item We propose a proxyless architecture adaptation algorithm that consecutively improves architecture and train the weights.
    \item We carried out extensive experiments on both supervised learning and self-supervised learning, and the results show consistent performance improvements.
    \item The proposed method overcomes the limitation of NAT such as poor reproducibility and limited types for the base architecture.
\end{itemize}

\begin{figure*}[t!]
\begin{center}
\centerline{\includegraphics[width=0.9\linewidth]{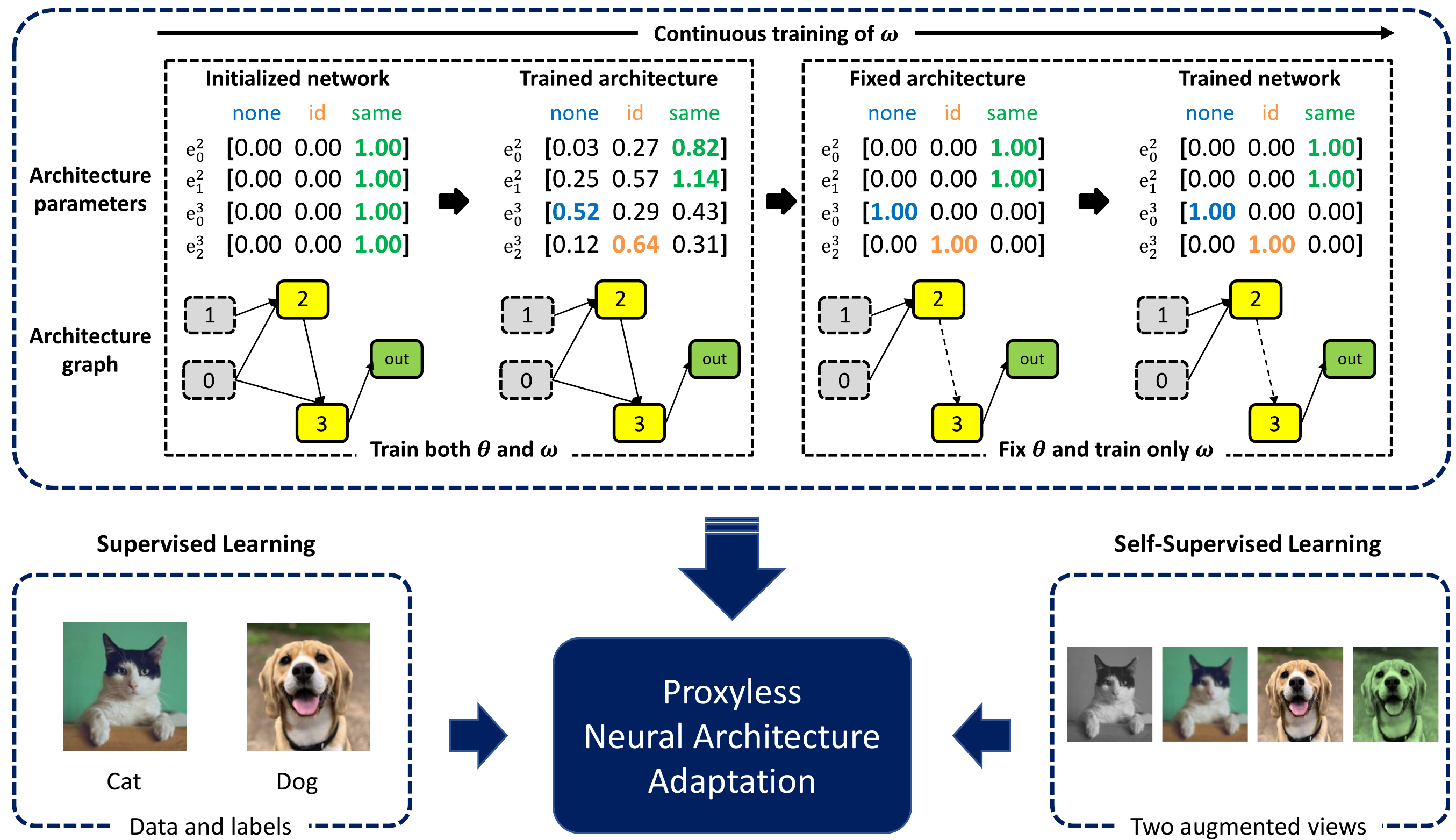}}
\caption{An example of the network architecture adaptation by the proposed method. Until the training of the architecture parameters $\theta$ is finished, both $\omega$ and $\theta$ are trained. After the architecture train step, $\theta$ is fixed and only $\omega$ is trained until we get the final trained network.}
\vspace{-0.5cm}
\label{overall}
\end{center}
\end{figure*}

\section{Related Work}
\label{Related}
After the NAS is introduced by~\cite{zoph2016neural}, various kinds of NAS methods have been proposed; reinforcement learning (RL)-based methods~\cite{zoph2018learning, pham2018efficient, lee2020efficient}, evolutionary algorithm (EA)-based method~\cite{real2017large, real2018regularized}, gradient-based~\cite{liu2018darts, luo2018neural}, and others~\cite{brock2017smash}.
Recently, NAT~\cite{guo2019nat} proposed the architecture adaptation that optimizes given neural architecture and search space.
Unlike traditional NAS methods, NAT only changes the original operations into none or identity operations.
Although NAT showed impressive results in the paper, there are several limitations of the NAT as we claimed in Section 1.

The proposed method can be applied to not only supervised learning but also self-supervised representation learning.
Traditionally, self-supervised learning focuses on learning representations from unlabeled data and pre-defined handcrafted tasks\cite{noroozi2016unsupervised, komodakis2018unsupervised}.
Recently suggested contrastive learning enables learning representations without defining specific tasks.
Especially, SimCLR~\cite{chen2020simple} achieved remarkable performance on various benchmarks with the utilization of effective augmentations on contrastive loss.

\section{Methodology}
\label{Method}

The proposed method adapts the given neural architecture and improves the performance by using gradient-based optimization.
After the entire learning process, we directly get the trained network, and there is no need to train a new network from scratch.
There are two consecutive stages in the proposed method: the architecture train stage and the network train stage.
In the architecture train stage, both the architecture parameters and network parameters are trained.
After the architecture train stage, only network parameters are trained.
The overall processes of the proposed method for both supervised learning and self-supervised learning are shown in Figure~\ref{overall}.

\subsection{Differentiable architecture parameters}
To adapt the given neural architecture, we use differentiable architecture parameters and gradient-based learning.
The architecture parameters $\theta$ is defined in the network architecture graph.
Each edge in the network architecture graph contains original operation, identity operation, and none operation.
Computation of each edge is carried out based on the architecture parameters:
\begin{equation}
\label{arch_params}
    {o}_{e}(x)={\theta}_{e, none}\cdot Z + {\theta}_{e, id}\cdot x + {\theta}_{e, same}\cdot o(x),
\end{equation}
where $x$ means input, ${o}_{e}(x)$ means output of the edge, $Z$ means zero tensor, $o(x)$ means original operation of the edge, ${\theta}_{e, none}$ means the weight of the none operation of the edge, ${\theta}_{e, id}$ means the weight of the identity operation of the edge, and ${\theta}_{e, same}$ means the weight of the original operation of the edge.
We set initial ${\theta}_{none}$ and ${\theta}_{id}$ as zero, and ${\theta}_{same}$ as one.
Therefore, the initialized network works the same as the original architecture.

\subsection{Architecture train stage}
After the network is initialized, the architecture train stage begins to adapt the neural architecture for the given dataset and task.
In this stage, both the network weight parameters $\omega$ and the architecture parameters $\theta$ are trained alternately.
For every input mini-batches, $\omega$ is trained first, and $\theta$ is trained after the update of $\omega$.
Note that the proposed method doesn't require any separated dataset for architecture optimization, and the network can utilize a full dataset to train its weights $\omega$.
The architecture train stage is carried out for the pre-defined epochs.

When the architecture train stage is finished, architecture is transformed based on the trained $\theta$.
For each edge, the operation that has the highest weight in $\theta$ is selected to construct the final architecture.
In the example of Figure~\ref{overall}, two edges maintain the original operations, one edge changed its operation into identity, and one edge is removed because none operation is selected.

\subsection{Network train stage}
The architecture trained at the previous stage is fixed, and only $\omega$ of the network is trained in this stage.
This stage is the same as the traditional neural network training process, and it is continuously carried out after the architecture train stage.
At the end of this stage, we can get the trained network and use it to infer unseen input data or test the performance of the model.

\subsection{Supervised learning and self-supervised learning}
The proposed method can be applied to both supervised learning and self-supervised learning.
The objective of the proposed method can be formulated as:

\begin{equation}
    \min_{\theta} \mathcal{L}\big(w^*(\theta), \theta\big)
\label{eq:gradient_obj}
\end{equation}
\begin{equation}
    \text{s.t.}~~~w^*(\theta) = \argmin_{w} \mathcal{L}(w, \theta),
\label{eq:gradient_w_obj}
\end{equation}

where $\theta$ denotes the architecture parameter, $w$ is the weight of the network, $\mathcal{L}(\cdot)$ is the loss function.

In case of the supervised learning, we used the cross-entropy loss which uses data and labels to calculate the loss.
In self-supervised learning experiments, we utilized SimCLR~\cite{chen2020simple} objective which uses two different augmented views of images to calculate the loss.

\section{Experiments}
\label{Experiments}
We carried out extensive experiments to verify the performance and the reproducibility of comparison methods.
In the experiments, various models are trained on CIFAR-10 and Tiny Imagenet.

\subsection{Data and Experiment Setting}
\textbf{Supervised Learning.}  CIFAR-10 dataset consists of 50,000 train images and 10,000 test images with ten classes.
The size of images is $32\times32$, and images have RGB color channels.
Tiny Imagenet dataset has 100,000 train images and 10,000 test images with 200 classes.
The input size of Tiny Imagenet dataset is $64\times64$, and all images are RGB color images.

We experimented with various models on CIFAR-10 and Tiny Imagenet dataset.
These models include ResNet20, MobileNet V2, DARTS, and ProxylessNAS.
The former two models are manually designed, and the latter two models are NAS models.
To compare the performance of NAT and our algorithm, we trained the NAT controller on each dataset and then trained the transformed architecture inferred from the controller.
In the case of our algorithm, we test both cell-based transformation and full network transformation.
%For comparison between NAT algorithm and our algorithm, NAT algorithm trained weights by searched architecture from the trained controller of NAT on each dataset, and our algorithms trained two ways that were trained from cells and the whole networks.
We used 0.025 learning rate, 600 epoch, and Stochastic Gradient Descent(SGD) optimizer as the same hyper-parameters to all models and methods.
Exceptionally, we applied 300 epochs to Mobilenet V2 and DARTS on Tiny Imagenet dataset, and utilized cut-out for NAS models such as DARTS and ProxylessNAS.

We tested five times with different random seeds for every experiment to get the right performance and verify the reproducibility of comparison algorithms.
Therefore, we report the average accuracy and standard deviation of each method and each model.
The total cost of NAT was calculated by adding GPU hours of the architecture transformation stage and network train stage, and the cost of our algorithms was computed by just measuring the cost of the whole training process.

\begin{table}[t]
\caption{Comparison of Average Accuracy, Standard Deviation and Total Cost between original, NAT and Ours on CIFAR-10.}
\label{tab_cifar10_result}
\vskip 0.05in
\begin{center}
% \begin{small}
\scalebox{0.8}{
\begin{tabular}{clccccc}
\toprule
& \multicolumn{1}{c}{}& \multicolumn{1}{c}{Avg Acc} & \multicolumn{1}{c}{Std} & \multicolumn{1}{c}{Total Cost}\\
Model & \multicolumn{1}{l}{Method} & \multicolumn{1}{c}{(\%)} & \multicolumn{1}{c}{(\%)}& \multicolumn{1}{c}{(GPU hours)}\\

\midrule
\multirow{4}{*}{Resnet20} & Original & 91.66 & 0.16 & 8\\
& NAT & 55.78 & 42.34 &  14\\
\cmidrule{2-5}
& Ours(Cell) & \textbf{93.29} & 0.11 & 11.2\\
& Ours(Full) & 93.12 & 0.12 & 8.7\\
\midrule
\multirow{4}{*}{\shortstack{Mobilenet \\ V2}} 
& Original & 93.91 & 0.12 & 20.5\\
& NAT & 91.97 & 5.10 & 27.8 \\
\cmidrule{2-5}
& Ours(Cell) & \textbf{95.02} & 0.31 & 24.8\\
& Ours(Full) & 94.93 & 0.13 & 22.1\\
\midrule
\midrule
\multirow{4}{*}{DARTS} & Original & 96.75 & 0.11 & 38.3\\
& NAT & 96.95 & 0.09 & 47 \\
\cmidrule{2-5}
& Ours(Cell) & \textbf{96.97} & 0.15 & 45\\
& Ours(Full) & 96.82 & 0.13 & 41.1\\
\midrule
\multirow{3}{*}{\shortstack{Proxyless \\ NAS}} & Original & 94.19 & 1.08 & 15.3\\
& NAT & - & - & - \\
\cmidrule{2-5}
& Ours(Full) & \textbf{95.09} & 0.23 & 19.8\\
\bottomrule
\end{tabular}
}
%\end{small}
\end{center}
\vskip -0.1in
\end{table}

\begin{table}[t]
\caption{Reproducibility of original, NAT and Ours with different random seeds on CIFAR-10.}
\label{tab_reproducibility}
\vskip 0.05in
\begin{center}
%\begin{small}
\scalebox{0.8}{
\begin{tabular}{clccccc}
\toprule
& \multicolumn{1}{c}{}& \multicolumn{5}{c}{Random Seed}\\
Model & \multicolumn{1}{l}{Method} & \multicolumn{1}{c}{(1)} & \multicolumn{1}{c}{(2)}& \multicolumn{1}{c}{(3)}& \multicolumn{1}{c}{(4)}& \multicolumn{1}{c}{(5)}\\

\midrule
\multirow{4}{*}{Resnet20} & Original & 91.74 & 91.74 & 91.65 & 91.76 & 91.39\\
& NAT & 10 & 75.14 & 91.05 & 10 & 92.68 \\
\cmidrule{2-7}
& Ours(Cell) & 93.21 & 93.34 & 93.4 & 93.15 & 93.37\\
& Ours(Full) & 93.07 & 93.26 & 93.22 & 93.06 & 92.97\\
\midrule
\multirow{4}{*}{\shortstack{Mobilenet \\ V2}} 
& Original & 93.9 & 94.04 & 93.95 & 93.95 & 93.72\\
& NAT & 83.36 & 95.13 & 94.95 & 91.18 & 95.21 \\
\cmidrule{2-7}
& Ours(Cell) & 94.57 & 94.97 & 95.41 & 94.97 & 95.18\\
& Ours(Full) & 95.13 & 94.81 & 94.85 & 95 & 94.85\\
\bottomrule
\end{tabular}
}
%\end{small}
\end{center}
\vskip -0.1in
\end{table}

\noindent\textbf{Self-Supervised Learning.} We conducted experiments with CIFAR-10 dataset to compare performances of original architectures and architectures adapted by the proposed method on self-supervised learning. Self-supervised learning requires deeper architecture than that used in supervised learning. Self-supervised learning is a kind of representation learning, which necessitates larger parameters to be converged. So we employed ResNet50 instead of ResNet20, and we adopted 24 
cells instead of 9 cells to DARTS architecture. As a result, the size of the parameters was 24.6$\pm$0.1 for self-supervised learning and the dimensions of the channels were adjusted to match the parameters. 
We used a temperature of 0.5, k of 200, epochs of 500, and 0.001 learning rate with Adam optimizer from SimCLR~\cite{chen2020simple} basic training settings. In all experiments, we maintained proxyless feature of the proposed method, which means we carried out the architecture search and learning the weights all at once.

\begin{figure}[t]
\begin{center}
\includegraphics[width=1.0\linewidth]{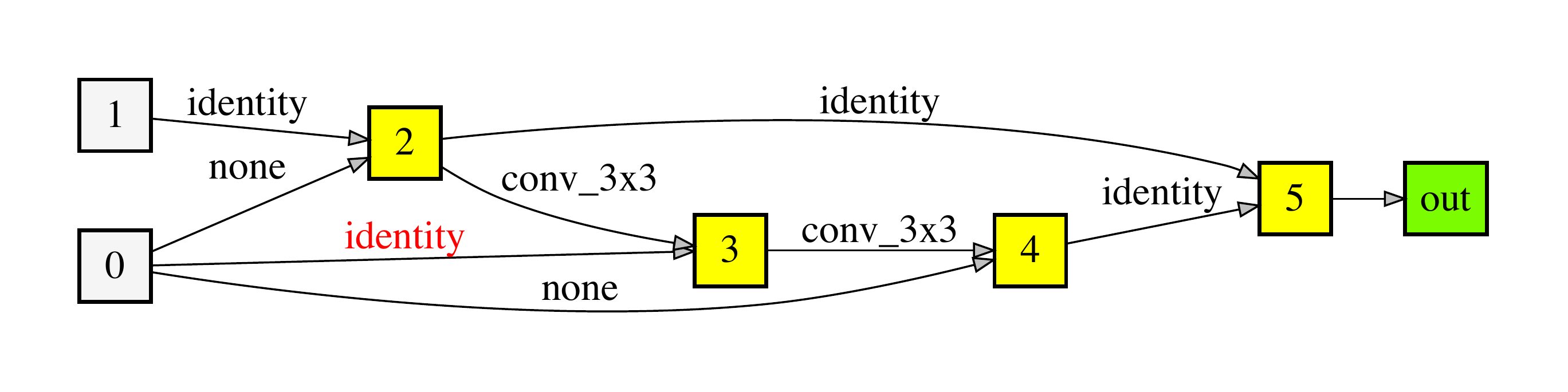}
\vspace{-0.2cm}
\centerline{\small (a) Ours}
\includegraphics[width=1.0\linewidth]{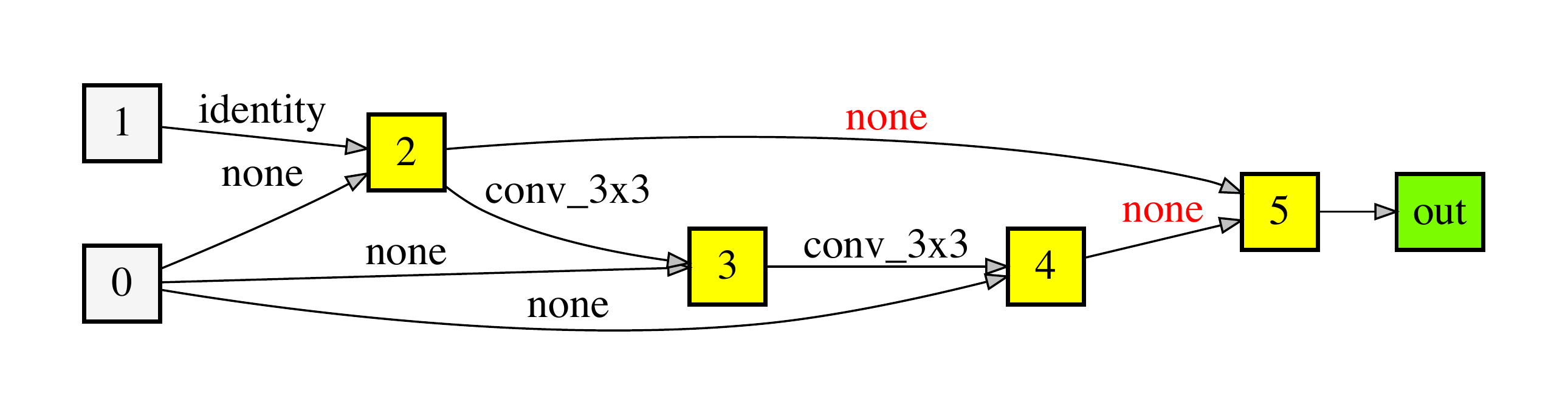}
\centerline{\small (b) NAT}
\vspace{-0.2cm}
\caption{Transformed ResNet20 cell architectures}
\vspace{-0.5cm}
\label{resnet}
\end{center}
\end{figure}

\begin{figure}[t]
\begin{center}
\includegraphics[width=1.0\linewidth]{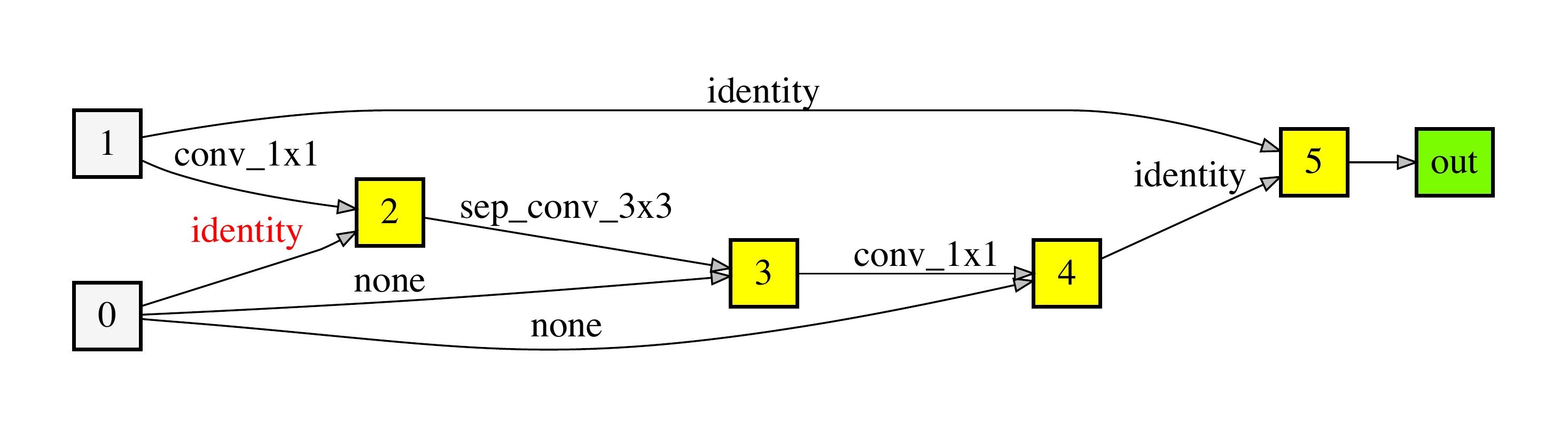}
\vspace{-0.2cm}
\centerline{\small (a) Ours}
\includegraphics[width=1.0\linewidth]{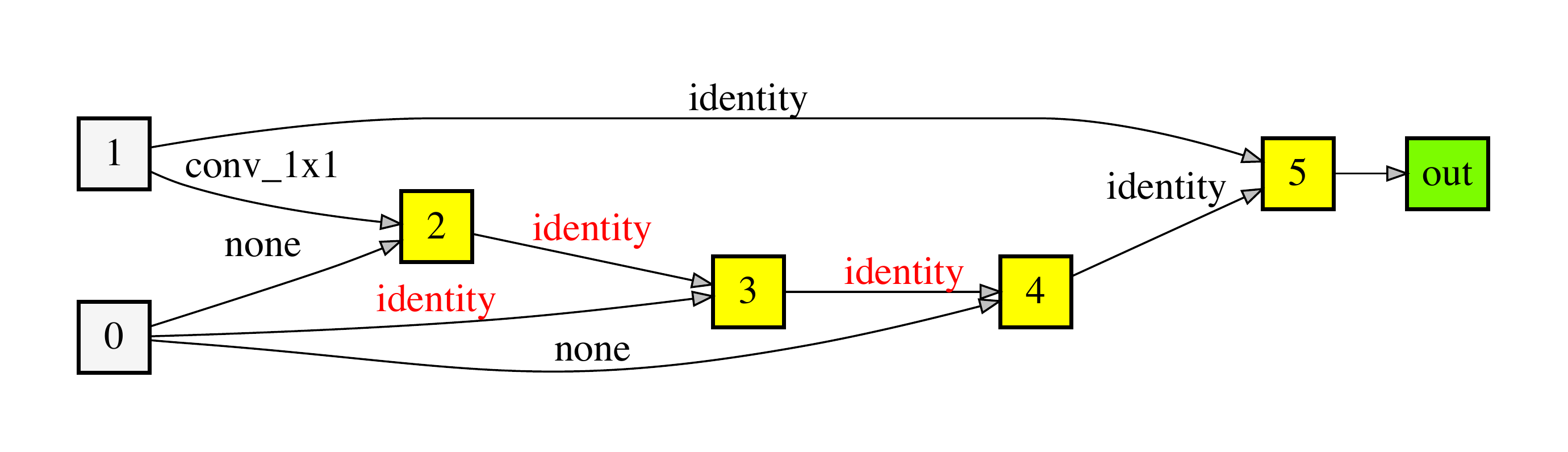}
\centerline{\small (b) NAT}
\vspace{-0.2cm}
\caption{Transformed MobileNetV2 cell architectures}
\vspace{-0.5cm}
\label{mobilenet}
\end{center}
\end{figure}

\begin{figure}[t]
\begin{center}
\includegraphics[width=0.7\linewidth]{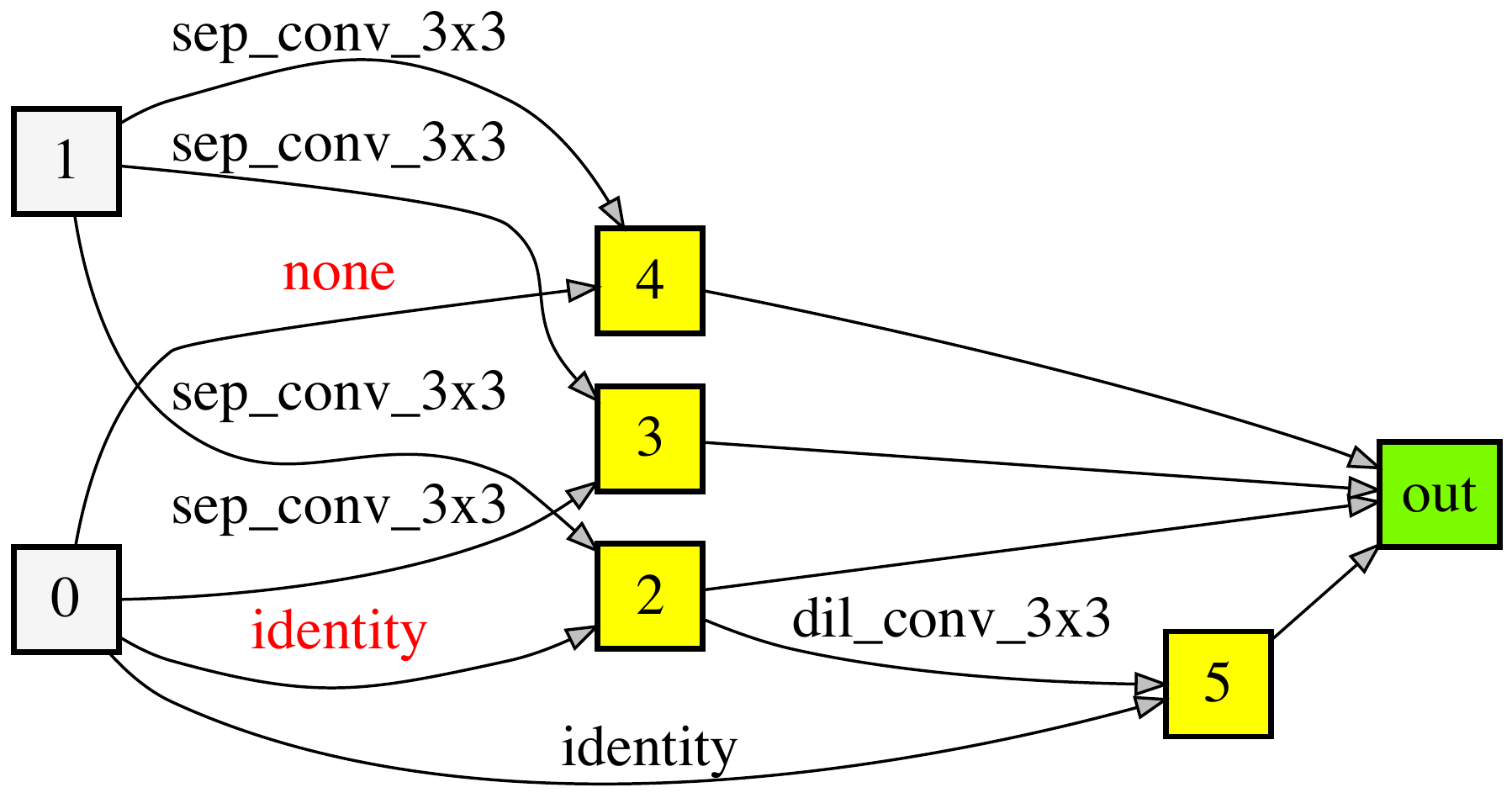}
% \vspace{0.1cm}
% \centerline{\small (a) Normal cell}
% \includegraphics[width=0.7\linewidth]{figures/reduction_darts_ours.pdf}
% \centerline{\small (b) Reduction cell}
% \vspace{-0.2cm}
\caption{DARTS normal cell transformed by our method. Reduction cell is the same as the original one.}
% \vspace{-0.5cm}
\label{darts}
\end{center}
\end{figure}

\subsection{Results on Supervised Learning}
The results of Table~\ref{tab_cifar10_result} have average accuracy, standard deviation, and total cost by various methods with different models on CIFAR-10 dataset.
We trained and inferred five times to get average accuracy and standard deviation.
As shown in Table~\ref{tab_cifar10_result}, the results of NAT is unstable in the case of manually designed models.
The results of our algorithms have better average accuracy and standard deviation than original and NAT in all cases. 
Moreover, the total computational cost is lower than NAT. 
Note that NAT cannot transform the architecture of ProxylessNAS, since it has various cell architectures in the network.
However, the proposed method successfully improves the performance of ProxylessNAS architecture.

Table~\ref{tab_reproducibility} shows the reproducibility of various methods with different random seeds on CIFAR-10 dataset.
Only one result is presented for each model in NAT paper. Therefore we experimented five times to get the right performance.
In the case of Resnet20 experiments, the results of seed 1 and 4 of NAT are caused by transforming identity edges into none operation.
Transformed Resnet20 architectures of seed 1 are presented in Figure~\ref{resnet}.
Changed edges are notated as red colors.
As shown in Figure~\ref{resnet}(b), NAT transformed all edges to node 5 into none operation.
Therefore, zero tensors are passed to the next layer.

Regarding the result of Mobilenet V2 experiments, the performance of NAT is degraded when it transforms convolution operation into identity operation.
Figure~\ref{mobilenet} shows the transformed Mobilenet V2 architectures of seed 1.
Additionally, we represent the transformed DARTS normal cell architecture of our algorithm in Figure~\ref{darts}.
There is no change in the edges of the reduction cell.

Table~\ref{tab_tiny_result} shows the results of various methods with different models on Tiny Imagenet dataset.
The results of this table describe that both our algorithms have better average accuracy and standard deviation than the original method upon all models. 

\begin{table}[t]
\caption{Comparison Average Accuracy, Standard Deviation and Total Cost between original, NAT and Ours on Tiny Imagenet.}
\label{tab_tiny_result}
\vskip 0.05in
\begin{center}
%\begin{small}
\scalebox{0.8}{
\begin{tabular}{clccccc}
\toprule
& \multicolumn{1}{c}{}& \multicolumn{1}{c}{Avg Acc} & \multicolumn{1}{c}{Std} & \multicolumn{1}{c}{Total Cost}\\
Model & \multicolumn{1}{l}{Method} & \multicolumn{1}{c}{(\%)} & \multicolumn{1}{c}{(\%)}& \multicolumn{1}{c}{(GPU hours)}\\

\midrule
\multirow{3}{*}{Resnet20} & Original & 50.72 & 0.41 & 16.1\\
\cmidrule{2-5}
& Ours(Cell) & \textbf{52.86} & 0.49 & 22\\
& Ours(Full) & 52.7118 & 0.23 & 17.5\\
\midrule
\multirow{3}{*}{\shortstack{Mobilenet \\ V2}} 
& Original & 51.57 & 0.76 & 20.3\\
\cmidrule{2-5}
& Ours(Cell) & \textbf{53.17} & 1.03 & 25.5\\
& Ours(Full) & 52.92 & 0.50 & 23.7\\
\midrule
\midrule
\multirow{3}{*}{DARTS} & Original & 59.25 & 0.44 & 39.2\\
\cmidrule{2-5}
& Ours(Cell) & 60.24 & 0.35 & 47\\
& Ours(Full) & \textbf{60.63} & 0.50 & 43.8\\
\bottomrule
\end{tabular}
}
%\end{small}
\end{center}
\vskip -0.1in
\end{table}

\subsection{Results on Self-Supervised Learning}
The results reported in Table~\ref{tab_selfsup_result} is evaluated by k-nearest neighbor (KNN) during the representation learning.
Therefore, it may not be better than the performance of linear evaluation reported in~\cite{chen2020simple} but we evaluated all architectures fair conditions.
% The evaluation method in Table~\ref{tab_selfsup_result} used k-nearest neighbor(KNN) on features extracted from the learned model, so it may not be better than using additional linear layer of updating weights one more for self-supervised learning.
According to the results of Table~\ref{tab_selfsup_result}, the proposed method outperforms the basic architecture of ResNet50 and DARTS even with similar parameter sizes.

In ResNet50, three operations were changed from none to skip-connection compared to the original architecture. And in DARTS, skip-connection was changed to none in only two places in the reduce cell, resulting in little difference in the size of parameters compared to the cases of supervised learning. 
This is due to the fact that self-supervised learning starves a lot of parameters than supervised learning.
% This is due to the fact that self-supervised learning requires more parameters than supervised learning.

\begin{table}[t]
\caption{Comparison Accuracy between original and Ours on Self-Supervised Learning.}
\label{tab_selfsup_result}
\vskip 0.05in
\begin{center}
%\begin{small}
\begin{tabular}{cclc}
\toprule
\multicolumn{1}{c}{}& \multicolumn{1}{c}{} & \multicolumn{1}{c}{} & \multicolumn{1}{c}{Accuracy}\\
DataSet & Model & \multicolumn{1}{l}{Method} & \multicolumn{1}{c}{(\%)}\\
\midrule
\multirow{5}{*}{CIFAR-10} & \multirow{2.5}{*}{Resnet50} & Original & 82.39\\
\cmidrule{3-4}
& & Ours & \textbf{86.8}\\
\cmidrule{2-4}
& \multirow{2.5}{*}{DARTS} & Original & 88.56\\
\cmidrule{3-4}
& & Ours & \textbf{89.04}\\
\bottomrule
\end{tabular}
%\end{small}
\end{center}
\vskip -0.1in
\end{table}

\section{Conclusion}
\label{Conclusion}

%We proposed a novel neural architecture transformation algorithm which being differentiable and reproducible for architecture improvement.
We proposed a novel gradient-based neural architecture adaptation algorithm that is reproducible and effective for architecture improvement.
%Since our algorithm utilized differentiable architecture parameters, we are able to train weights at once, including non-cell structure.
%Thanks to the differentiable architecture parameters, our algorithm can transform the architecture and train the network consecutively.
Thanks to the differentiable architecture parameters, our algorithm can train both the architecture and the network at once.
%Also, both cell architectures and full network architectures can be improved by the proposed method.
%The results of five times experiments of all methods and models demonstrate that the proposed algorithm has high reproducibility and better performance than the original and NAT method on CIFAR-10 and Tiny Imagenet dataset. 
The results of the experiments demonstrate that the proposed algorithm has high reproducibility and stably improves the performance of various models on various datasets.
Moreover, the proposed method can be applied to both supervised learning and self-supervised learning and achieve performance improvement on both learning schemes.

{\small
\bibliography{ref}

\begin{thebibliography}{13}
\providecommand{\natexlab}[1]{#1}

\bibitem[{Brock~et al.(2017)}]{brock2017smash}
Brock~et al., A. 2017.
\newblock SMASH: one-shot model architecture search through hypernetworks.
\newblock \emph{arXiv preprint arXiv:1708.05344}.

\bibitem[{Chen~et al.(2020)}]{chen2020simple}
Chen~et al., T. 2020.
\newblock A simple framework for contrastive learning of visual
  representations.
\newblock In \emph{ICML}, 1597--1607. PMLR.

\bibitem[{Guo et~al.(2019)Guo, Zheng, Tan, Chen, Chen, Zhao, and
  Huang}]{guo2019nat}
Guo, Y.; Zheng, Y.; Tan, M.; Chen, Q.; Chen, J.; Zhao, P.; and Huang, J. 2019.
\newblock NAT: Neural Architecture Transformer for Accurate and Compact
  Architectures.
\newblock In \emph{Advances in Neural Information Processing Systems},
  735--747.

\bibitem[{Komodakis and Gidaris(2018)}]{komodakis2018unsupervised}
Komodakis, N.; and Gidaris, S. 2018.
\newblock Unsupervised representation learning by predicting image rotations.
\newblock In \emph{International Conference on Learning Representations}.

\bibitem[{Lee, Kim, and Han(2020)}]{lee2020efficient}
Lee, H.-C.; Kim, D.-G.; and Han, B. 2020.
\newblock Efficient decoupled neural architecture search by structure and
  operation sampling.
\newblock In \emph{2020 IEEE ICASSP}, 4222--4226. IEEE.

\bibitem[{Liu, Simonyan, and Yang(2019)}]{liu2018darts}
Liu, H.; Simonyan, K.; and Yang, Y. 2019.
\newblock Darts: Differentiable architecture search.
\newblock \emph{Proceedings of ICLR}.

\bibitem[{Luo et~al.(2018)Luo, Tian, Qin, Chen, and Liu}]{luo2018neural}
Luo, R.; Tian, F.; Qin, T.; Chen, E.; and Liu, T.-Y. 2018.
\newblock Neural architecture optimization.
\newblock In \emph{Advances in neural information processing systems},
  7816--7827.

\bibitem[{Noroozi and Favaro(2016)}]{noroozi2016unsupervised}
Noroozi, M.; and Favaro, P. 2016.
\newblock Unsupervised learning of visual representations by solving jigsaw
  puzzles.
\newblock In \emph{ECCV}, 69--84. Springer.

\bibitem[{Pham et~al.(2018)Pham, Guan, Zoph, Le, and Dean}]{pham2018efficient}
Pham, H.; Guan, M.~Y.; Zoph, B.; Le, Q.~V.; and Dean, J. 2018.
\newblock Efficient neural architecture search via parameter sharing.
\newblock \emph{Proceedings of ICML}.

\bibitem[{Real~et al.(2017)}]{real2017large}
Real~et al., E. 2017.
\newblock Large-scale evolution of image classifiers.
\newblock \emph{arXiv preprint arXiv:1703.01041}.

\bibitem[{Real~et al.(2018)}]{real2018regularized}
Real~et al., E. 2018.
\newblock Regularized evolution for image classifier architecture search.
\newblock \emph{arXiv preprint arXiv:1802.01548}.

\bibitem[{Zoph and Le(2017)}]{zoph2016neural}
Zoph, B.; and Le, Q.~V. 2017.
\newblock Neural architecture search with reinforcement learning.
\newblock \emph{Proceedings of ICLR}.

\bibitem[{Zoph~et al.(2018)}]{zoph2018learning}
Zoph~et al., B. 2018.
\newblock Learning transferable architectures for scalable image recognition.
\newblock In \emph{Proceedings of the IEEE conference on CVPR}, 8697--8710.

\end{thebibliography}
}

\end{document}